\documentclass[a4paper,fleqn]{cas-dc}
\usepackage[authoryear,round]{natbib}
\usepackage{caption}

\usepackage{multirow}

\usepackage{microtype}

\usepackage{hyperref}

\usepackage{booktabs}
\usepackage{multirow}
\usepackage{xcolor}
\def\tsc#1{\csdef{#1}{\textsc{\lowercase{#1}}\xspace}}
\tsc{WGM}
\tsc{QE}

\begin{document}
\let\WriteBookmarks\relax
\def\floatpagepagefraction{1}
\def\textpagefraction{.001}
\let\printorcid\relax 

\newcommand{\tabfo}{\fontsize{10pt}{12pt}\selectfont} 
\newcommand{\tabfoa}{\fontsize{9pt}{10pt}\selectfont}
\newcommand{\tabfob}{\fontsize{8pt}{9pt}\selectfont}
\newcommand{\tabfont}{\fontsize{6.7pt}{9pt}\selectfont}
\newcommand{\tabf}{\fontsize{5.7pt}{8pt}\selectfont}

\shorttitle{}    

\shortauthors{Junyao Li et al.}

\title[mode = title]{Exploiting Gaussian Agnostic Representation Learning with Diffusion Priors for Enhanced Infrared Small Target Detection}  

\author[1]{Junyao Li}
\author[1]{Yahao Lu}
\author[1]{Xingyuan Guo}
\author[2]{Xiaoyu Xian}
\author[3]{Tiantian Wang}
\author[1]{Yukai Shi}
\ead{<ykshi@gdut.edu.cn>}
\cormark[1]

\address[1]{School of Information Engineering, Guangdong University of Technology, Guangzhou, 510006, China} 
\address[2]{CRRC Insitution, Beijing 100000, China} 
\address[3]{Guangzhou National Laboratory, Guangzhou 510006, China} 
\cortext[1]{Corresponding author}  

\begin{abstract}
Infrared small target detection (ISTD) plays a vital role in numerous practical applications. In pursuit of determining the performance boundaries, researchers employ large and expensive manual-labeling data for representation learning. Nevertheless, this approach renders the state-of-the-art ISTD methods highly fragile in real-world challenges. In this paper, we first study the variation in detection performance across several mainstream methods under various scarcity -- namely, the absence of high-quality infrared data -- that challenge the prevailing theories about practical ISTD. To address this concern, we introduce the Gaussian Agnostic Representation Learning. Specifically, we propose the Gaussian Group Squeezer, leveraging Gaussian sampling and compression for non-uniform quantization. By exploiting a diverse array of training samples, we enhance the resilience of ISTD models against various challenges. Then, we introduce two-stage diffusion models for real-world reconstruction. By aligning quantized signals closely with real-world distributions, we significantly elevate the quality and fidelity of the synthetic samples. Comparative evaluations against state-of-the-art detection methods in various scarcity scenarios demonstrate the efficacy of the proposed approach.
\end{abstract}







\maketitle

\section{Introduction}

Infrared small target detection (ISTD)  \citep{zhang2025m4net, shen2025graph, Alpher01} plays a vital role in numerous practical applications, particularly in the fields of video surveillance \citep{xiao2023ediffsr,hu2023cycmunet,xiao2304local}, early warning systems \citep{deng2016small} and environmental monitoring \citep{rawat2020review}. To effectively detect small targets in infrared images, researchers have proposed a variety of filter-based and neural network-based methods for processing real-world data~\cite{lu2024sirst,shi2024diff}. In contrast to traditional methods, neural network-based \citep{Deeplearning} approaches learn the representations of infrared small targets from the \emph{expensive and numerous} manual labeling, thereby enabling better localization of these targets. 

Deep learning methodologies typically rely on such extensive, large-scale training datasets. However, acquiring an adequate quantity of high-quality labeled data remains a significant challenge. In the context of marine monitoring \citep{Teutsch2010,Ying2022}, small targets, such as long-range vessels or drones, often exhibit low contrast against intricate background environments. In complex environments—such as ocean waves, vegetation, and cloud formations—noise can readily obscure small targets \citep{wang2022mpanet}, further complicating their detection and accurate labeling. Moreover, the detection of small targets is frequently hindered by their considerable distance and the limited resolution of imaging sensors. Collectively, these physical challenges significantly impede data accessibility.

\begin{figure}[t]
  \centering
  \includegraphics[width=0.95\linewidth]{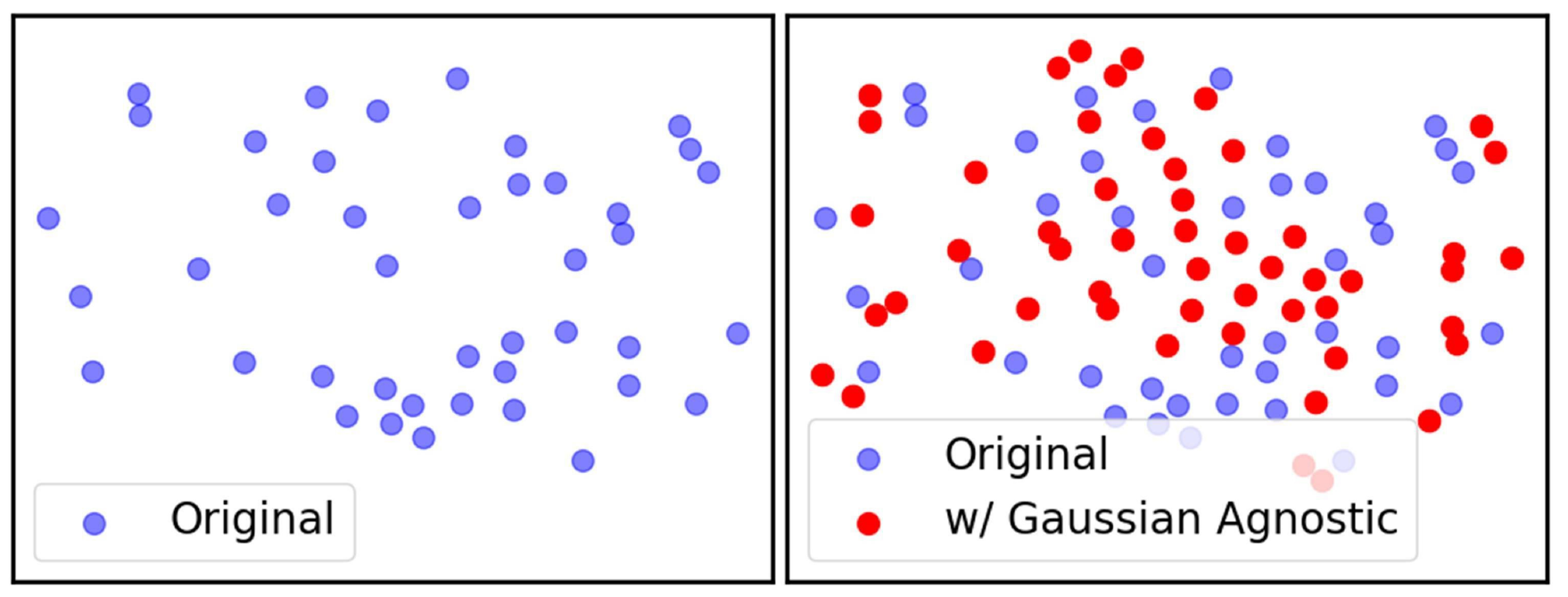}
   \caption{t-SNE \citep{vanderMaaten2008} visualizations of the infrared image features. The \textcolor{blue}{blue} points represent samples from the original SIRST \citep{Dai2021} dataset, while the \textcolor{red}{red} points represent the samples augmented by Gaussian Agnostic Representation Learning. It is evident that the augmented red points help to generate richer, compact and diverse feature points compared to the original blue samples.
}
   \label{fig:fig1}
   \vspace{-5mm}
\end{figure}

Hence, ISTD techniques are particularly crucial in scenarios characterized by data scarcity. However, most existing ISTD models exhibit suboptimal performance under extreme conditions. In real-world scenes, obtaining enough and high-precision data is unlikely to be achieved. Insufficient data prevents these models from effectively learning the essential target features, resulting in a marked decrease in detection accuracy. Moreover, due to complex and dynamic backgrounds combined with very low target contrast, extracting stable and discriminative features from limited samples becomes exceedingly challenging. Consequently, the robustness and accuracy of existing methods are significantly constrained under few-shot learning scenarios. To quantitatively assess the impact of data, we utilized 30\% and 10\% of the training data from the SIRST \citep{Dai2021} dataset to simulate training scenarios. \autoref{tab:table1} illustrates the variation in detection performance across several mainstream methods under both full-scale and few-shot scenarios. These results quantitatively indicate that the model can effectively learn features and exhibit superior performance with sufficient data. In contrast, existing methods demonstrate poor generalization ability and unstable performance in few-sample scenarios. As shown in \autoref{tab:table1}, sometimes, the state-of-the-art models struggle to learn sufficient feature representations, leading to a drastic decrease in detection accuracy.
\begin{table}[]\rmfamily
  \centering
  \tabfont
  \setlength{\tabcolsep}{2pt}
  \caption{To modeling various real-world cases, the training data is scaled from $100\%$ to $10\%$ setting on SIRST \citep{Dai2021} dataset.}
  \label{tab:performance_comparison}
  \begin{tabular}{cccccc}
    \toprule
    \multirow{2}{*}{\begin{tabular}[c]{@{}c@{}}Perfor-\\ mance\end{tabular}} & \multirow{2}{*}{\begin{tabular}[c]{@{}c@{}}Data \\ Ratio\end{tabular}} & \multicolumn{4}{c}{Method} \\
    \cmidrule{3-6}
                             & & ACM & DNANet & SCTNet & Ours \\
    \midrule
    \multirow{3}{*}{$IoU$}       & 100\%                       & 66.14 & 76.48 & 76.00 & \textbf{78.88} \\
                             & 30\%                        &
                             62.91(\textcolor{blue}{$\downarrow3.23\%$})\ & 74.56(\textcolor{blue}{$\downarrow1.92\%$}) & 69.42(\textcolor{blue}{$\downarrow6.58\%$}) & \textbf{78.19}(\textcolor{red}{$\downarrow0.69\%$}) \\
                             & 10\%                        & 20.84(\textcolor{blue}{$\downarrow45.30\%$}) & 56.49(\textcolor{blue}{$\downarrow19.99\%$}) & 59.74(\textcolor{blue}{$\downarrow16.26\%$}) & \textbf{67.96}(\textcolor{red}{$\downarrow10.92\%$}) \\
    \midrule
    \multirow{3}{*}{$P_d$}       & 100\%                       & 91.63 & 96.57 & 95.81 & \textbf{96.19} \\
                             & 30\%                        & 91.25(\textcolor{blue}{$\downarrow0.38\%$}) & 93.15(\textcolor{blue}{$\downarrow3.42\%$}) & 95.43(\textcolor{blue}{$\downarrow0.38\%$}) & \textbf{96.19}(\textcolor{red}{$\downarrow0\%$}) \\
                             & 10\%                        & 85.17(\textcolor{blue}{$\downarrow6.46\%$}) & 88.97(\textcolor{blue}{$\downarrow7.6\%$}) & 90.87(\textcolor{blue}{$\downarrow4.94\%$}) & \textbf{92.77}(\textcolor{red}{$\downarrow3.42\%$}) \\
    \bottomrule
  \end{tabular}
  \label{tab:table1}
  \vspace{-5mm}
\end{table}

To make ISTD models robust to data scarcity, we propose the Gaussian Agnostic Representation Learning. This approach aims to augment the available data for infrared small targets and enhance the performance of detection models toward various challenges. Specifically, we propose the Gaussian Group Squeezer, which is based on Gaussian sampling and compression. This module performs non-uniform quantization operations to generate an extensive number of training samples for diffusion models. Diffusion models comprise two stages: the coarse reconstruction stage and the diffusion stage. Meanwhile, the coarse-rebuilding stage learns the mapping representation from the quantized images to a reconstructed image, facilitating the initial pixel restoration. Additionally, we train the diffusion stage to resample the reconstructed data, thereby further aligning it with real-world distributions. We aim to expand the IR dataset by generating high-quality synthetic data using trained generative models. As illustrated in \autoref{fig:fig1}, the synthetic samples exhibit a more compact feature distribution compared to the blue dots representing the original SIRST \citep{Dai2021} samples. The generated data is able to be integrated into an arbitrary model and bring enhanced ISTD methods.  \autoref{tab:performance_comparison} demonstrates the effectiveness of the proposed enhancement strategy in improving representation learning and detection performance. Against extreme challenges, our approach significantly improves detection performance in scenarios ranging from few-shot to full-scale. The main contributions are as follows:
\begin{itemize}
\item[$\bullet$]We introduce Gaussian Agnostic Representation Learning, tailored to address data scarcity in Infrared Small Target Detection (ISTD) models. This advanced technique employs a Gaussian Group Squeezer, leveraging Gaussian sampling and compression for non-uniform quantization. By generating a diverse array of training samples, we enhance the resilience of ISTD models against various challenges, and expanding the infrared dataset with high-quality synthetic data.
\item[$\bullet$] Two-stage generative models for real-world reconstruction. The generative models learns to align quantized images closely with real-world distributions, facilitating precise pixel resampling. This integrated process significantly elevates the quality and fidelity of the synthetic samples.
\item[$\bullet$] Comparative evaluations against state-of-the-art detection methods in both full-scale and few-shot scenarios demonstrate the efficacy of the proposed approach.
\end{itemize}

\section{Related work}

\subsection{Single-frame Infrared Small Target Detection}
\begin{figure*}[ht]\rmfamily
  \centering
  \includegraphics[width=\textwidth]{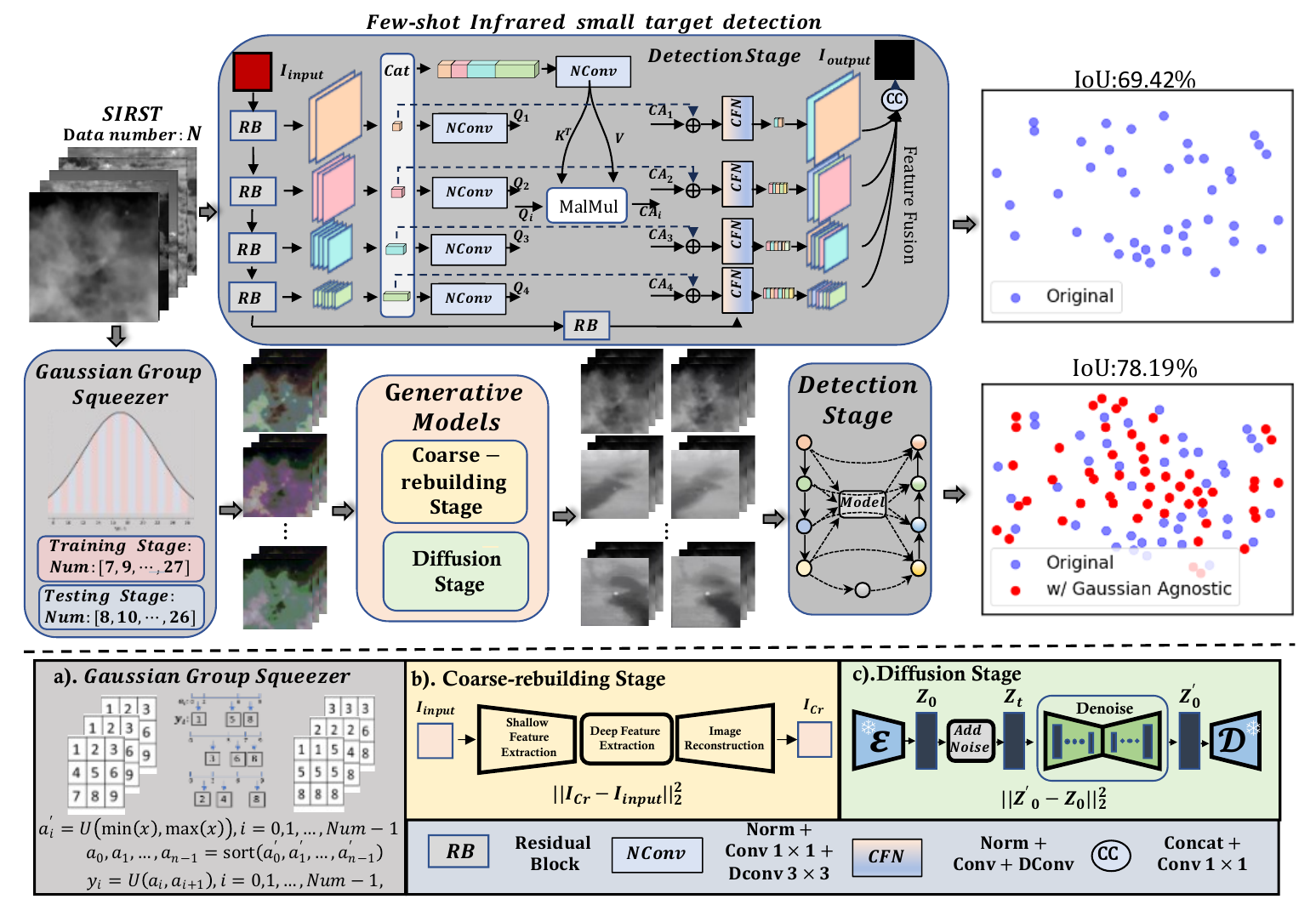}
   \caption{Gaussian Agnostic Representation Learning. Our approach generates substantial training data using a Gaussian-Agnostic strategy. By applying Gaussian Agnostic representation learning with Generative Models, the performance of ISTD is obviously enhanced, achieving higher IoU scores among various of extreme challenges. \emph{In training stage}, the framework comprises three stages: a) Gaussian Group Squeezer: sampling parameters from Gaussian distributions for quantizing images. b) Coarse-rebuilding Stage: fusing shallow and deep features to reconstruct quantized content initially. c) Diffusion Stage: hidden variables are encoded and decoded using VQ-VAE \citep{vq-vae} to generate realistic content through diffusion prior. \emph{In inference stage}, we shuffle the Gaussian parameterized interval to enforce the diffusion model generate diverse yet uniform data. The feature distribution of the generated samples (\textcolor{red}{red dots}) is more compact than that of the original samples (\textcolor{blue}{blue dots}), which facilitate a generalizable representation learning.
}
   \label{fig:fig2}
   \vspace{-5mm}
\end{figure*}

Single-frame infrared small target detection Compared to generic object detection,
infrared small target detection has several unique characteristics:\textbf{1) }\textbf{varying background clutter: }Clouds, building edges, and tree lines that generate high-frequency noise. Dynamic heat signatures from roads, machinery, or ocean waves \textbf{2) }\textbf{Low Target-Background Contrast: }Scattering and absorption over long distances. Small targets emit limited heat \citep{li2022dense} . \textbf{3) }\textbf{small target size: }Due to the long imaging distance, infrared
targets are generally small, ranging from one pixel to tens of pixels in the images \citep{li2022dense} .

ISTD for accurate segmentation~\cite{peng2025moving,yang2025istd} in complex backgrounds remains a challenging task. Infrared imaging is highly susceptible to noise interference, small target sizes, and extended imaging distances, which significantly increase the difficulty of detection. Traditional methods, such as Top-hat \citep{rivest1996detection}, LCM \citep{chen2013local}, and IPI models \citep{gao2013infrared}, struggle to effectively separate small targets from complex backgrounds due to the lack of semantic information utilization.

In contrast, CNN-based models~\cite{kumar2025small,zhong2024hierarchical,zhang2025m4net} have the capability to extract advanced semantic features, enabling more effective recognition of infrared small targets. MDvsFA-GAN \citep{wang2019miss} performs multimodal data transformation using generative adversarial networks. ISTDU-Net \citep{Xi} enhances small target features via feature map grouping. ACM \citep{dai2021asymmetric} combines global context with local details to improve detection accuracy. DNA-Net \citep{li2022dense} utilizes dense region nesting and attention mechanisms to capture multi-scale features. UIU-Net \citep{wu2022uiu} integrates both local and global information. ISNet \citep{Zhang2022} emphasizes target shape features to improve detection accuracy. MSHNet \citep{liu2024infrared} is suitable for scenarios with substantial target size variations. SpirDet \citep{mao2024spirdet} optimizes infrared small target detection with efficient and lightweight design. SCAFNet \citep{zhang2024scafnet} improves target recognition accuracy through semantically guided fusion. and SCTransNet \citep{10486932} leverages spatial and channel attention mechanisms to enhance detection performance. Despite the significant advancements of these CNN-based methods, the high costs of acquiring infrared image data, the cumbersome labeling process, and the limited data availability remain major bottlenecks hindering further performance improvements.

\begin{table}[t]\rmfamily
\centering
\caption{Comparison of diffusion-based ISTD methods under 30\% few-shot settings on the NUDT-SIRST dataset. Our method adopts diffusion for data augmentation and shows superior performance.}
\resizebox{\linewidth}{!}{ 
\begin{tabular}{lcccccc}
\toprule
\textbf{Method} & \textbf{Role of Diffusion}  & \textbf{Architecture} 
& \textbf{IoU} & \textbf{$P_d$} & \textbf{$F_a$}  \\
\midrule
ISTD-Diff     & Detection                 & Attn + Diffusion       & -     & -     & -     \\
Diff-Mosaic   & Data Augmentation        & Mosaic + Diffusion     & 89.82 & 98.41 & 12.70 \\
\textbf{Ours} & Data Augmentation        & GGS + 2-Stage Diff     & \textbf{93.39} & \textbf{98.94} & \textbf{2.06} \\
\bottomrule
\end{tabular}
}
\label{tab:fewshot_diffusion_compare}
\vspace{-2mm}
\end{table}

\subsection{Diffusion Models}
In recent years, Denoising Diffusion Probabilistic Models (DDPM) \citep{ho2020denoising} have achieved significant breakthroughs in the field of image generation \citep{lin2023diffbir,chen2023controlstyle}. The success of DDPM has significantly advanced the development of high-quality, photorealistic image generation techniques. 
To further enhance performance, researchers proposed the Latent Diffusion Model (LDM) \citep{rombach2022high}. LDM not only generates high-quality images but also incorporates realistic prior knowledge, demonstrating its powerful capabilities across multiple application scenarios. By compressing data into a low-dimensional latent space through an autoencoder, LDM significantly improves both training efficiency and generation quality. ControlStyle \citep{chen2023controlstyle} combines text and image data to achieve text-driven high-quality stylized image generation via diffusion priors. These applications illustrate that diffusion-based generation techniques have substantial potential and broad applicability in image processing and creative content generation.

Recently, several researchers have introduced powerful diffusion models into ISTD tasks to enhance model performance. ISTD-diff \citep{du2024ISTDdiff} frames the ISTD task as a generative task by iteratively generating target masks from noise. However, this approach has only led to limited improvements in detection performance and has increased computational complexity, resulting in reduced inference time. Diff-mosaic \citep{shi2024diff} method aims to effectively address the challenges of data augmentation methods in terms of diversity and fidelity by leveraging a diffusion prior. However, in few-shot conditions, the lack of sufficient data and effective data augmentation techniques to fine-tune the diffusion model can lead to model overfitting and subsequent performance degradation.

To validate the effectiveness of our method in few-shot scenarios, we evaluate all competing approaches using only 30\% of the NUDT-SIRST training data under the same training pipeline.
As shown in \autoref{tab:fewshot_diffusion_compare}, our method achieves the highest Intersection over Union (IoU) and probability of detection ($P_d$), while significantly reducing the false alarm rate ($F_a$) compared to Diff-Mosaic.

\section{Methodology}
\subsection{Problem Definition}

\begin{table}[t]\rmfamily
  \centering
  \setlength{\tabcolsep}{1pt}
  \caption{Main characteristics of the SIRST \citep{Dai2021} and NUDT-SIRST \citep{li2022dense} Datasets}
  \begin{tabular}{ccccc}
    \toprule
    \multirow{2}{*}{Datasets} & \multicolumn{3}{c}{Target size percent of image area} & \multirow{2}{*}{SCR} \\
    \cmidrule(lr){2-4}
    & 0$\sim$0.03\% & 0.03$\sim$0.15\% & \textgreater{}0.15\% & \\
    \midrule
    SIRST     & 58\%(percent of images) & 35\% & 7\%  & 6.24 \\
    NUDT-SIRST & 58\% & 38\% & 4\%  & 1.83 \\
    \bottomrule
  \end{tabular}
  \label{tab:dataset_chars}
  \vspace{-5mm}
\end{table}
\begin{figure}[h]
  \centering
  \includegraphics[width=0.95\linewidth]{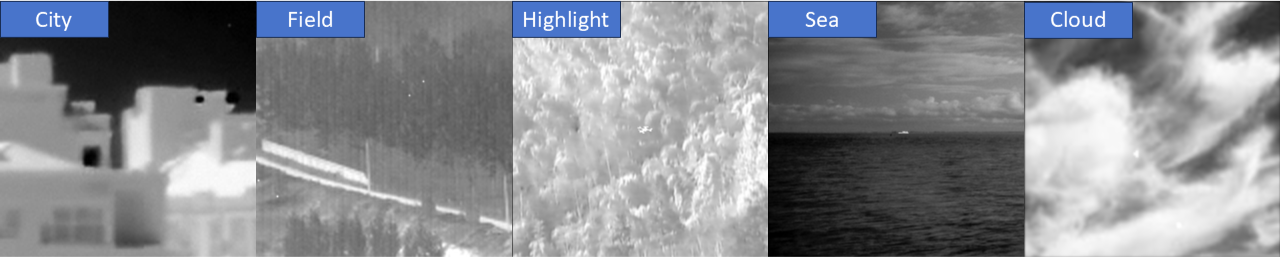}
   \caption{Real-world infrared scenes exhibit severe non-uniform backgrounds.}
   \label{fig:new_add_fig1}
   \vspace{-5mm}
\end{figure}

Infrared small target detection (ISTD) confronts three fundamental challenges: heterogeneous background clutter, extremely low target-to-background contrast, and sub-pixel target scales. As evidenced in \autoref{tab:dataset_chars}, over 90\% of targets occupy less than 0.15\% of the image area across SIRST \citep{Dai2021} and NUDT-SIRST datasets. Critical signal-to-clutter ratio(SCR) metrics further quantify contrast degradation: SIRST \citep{Dai2021}  averages 6.24 while NUDT-SIRST drops to 1.83 (approaching the detection boundary). \autoref{fig:new_add_fig1} visually corroborates these issues, where complex backgrounds (e.g., cloud clusters, sea glint) exhibit high-intensity variations that obscure sub-pixel targets.

ISTD methods typically rely on large-scale, high-quality training datasets. However, their performance in real-world scenarios often degrades significantly. To quantitatively assess this effect, we conducted experiments on the SIRST \citep{Dai2021} dataset, utilizing 30\% and 10\% of the training data to simulate each extreme condition. With 30\% training data, the IoU of SCTansNet \citep{10486932} drops from 76.00\% to 69.42\%, DNANet \citep{li2022dense} decreases from 76.48\% to 74.56\%, and ACM \citep{dai2021asymmetric} declines from 66.14\% to 62.91\%. At 10\% training data, SCTansNet \citep{10486932} further decreases to 59.74\%, DNANet \citep{li2022dense} to 56.49\%, and ACM \citep{dai2021asymmetric} to 20.84\%. The results indicate that high-quality training data enables the model to learn features more effectively. Conversely, with fewer and samples, the model's generalization ability easily be violated, leading to a fragile performance in ISTD.

\subsection{Gaussian Agnostic Representation Learning} \label{NA}
We propose a Gaussian-Agnostic Representation Learning method based on generative models to address the performance bottleneck in ISTD under various extreme challenges. Our framework comprises three modules: Gaussian group squeezer, coarse-rebuilding stage and diffusion stage. 

\begin{figure*}[h]
  \centering
  \includegraphics[width=0.95\linewidth]{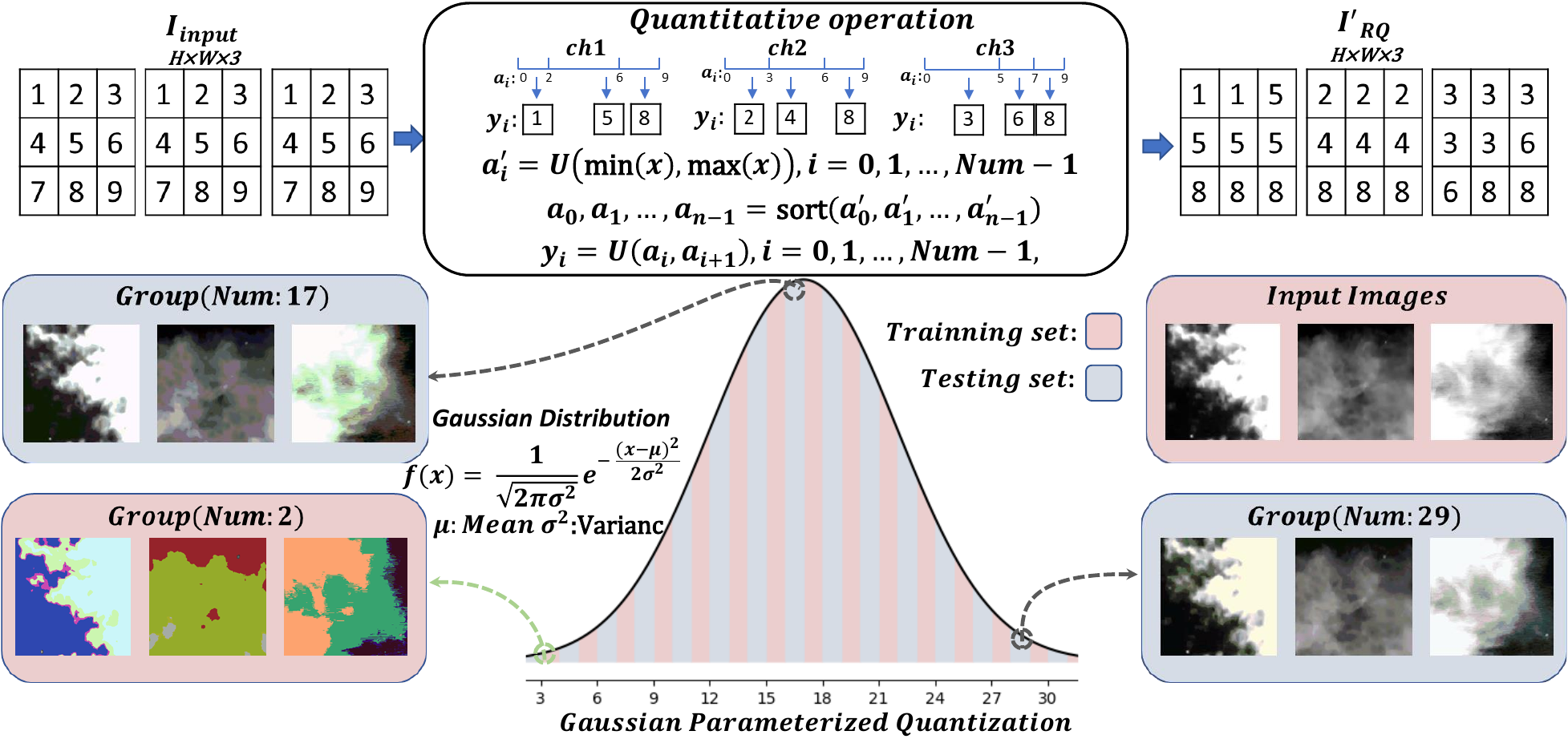}
   \caption{Diversification of infrared data with Gaussian group squeezer. We designed a non-uniform quantizer with randomized Gaussian interval sizes to implement the cross-peak sampling strategy. The pixel values of the input image are initially sampled and sorted. Next, random values are selected from each neighboring interval $U(a_i, a_{i+1})$ and used as replicated values $y_i$, replacing the pixels with their corresponding $y_i$ to generate the quantized image. $Num$ is sampled from a Gaussian distribution. We apply \emph{distinct} quantization parameters as inputs for generative models during the training and testing phases.
}
   \label{fig:fig3}
   \vspace{-5mm}
\end{figure*}

\begin{figure}[ht]
  \centering
  \includegraphics[width=0.95\linewidth]{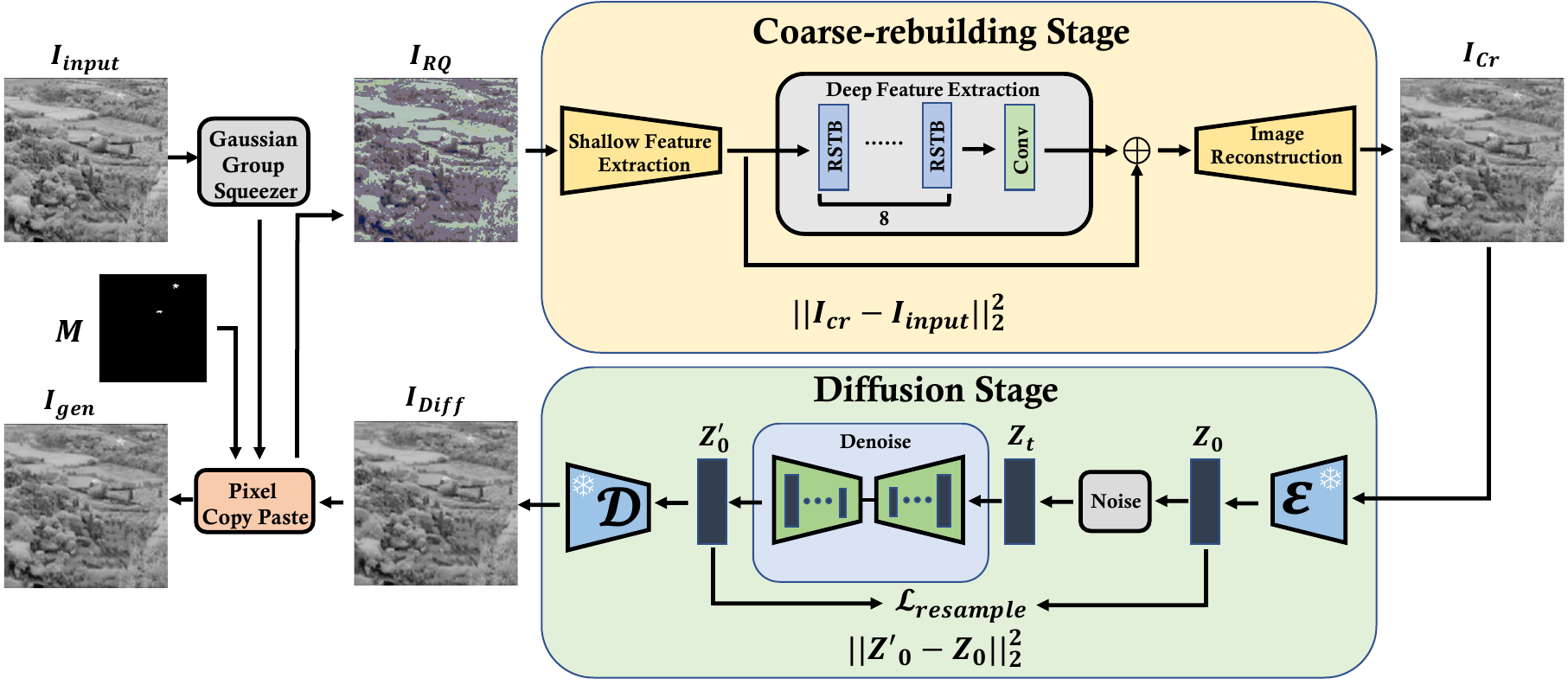}
   \caption{
     Two-stage Generative Models for Agnostic Representation Learning. The quantized image $I_{RQ}$ first passes through the coarse-rebuilding stage to repair quantization-corrupted pixels, generating a preliminary restored image $I_{Cr}$. Subsequently, $I_{Cr}$ enters the diffusion stage, where it learns the high-quality image distribution through the diffusion process, resulting in the generation of the synthesized image $I_{Diff}$. Finally, the small target pixels are copied into the generated image using the Pixel Copy Paste module to obtain the final synthesized image $I_{gen}$.
}
   \label{fig:fig5}
    \vspace{-5mm}
\end{figure}

\begin{figure}[ht]
  \centering
  \includegraphics[width=0.85\linewidth]{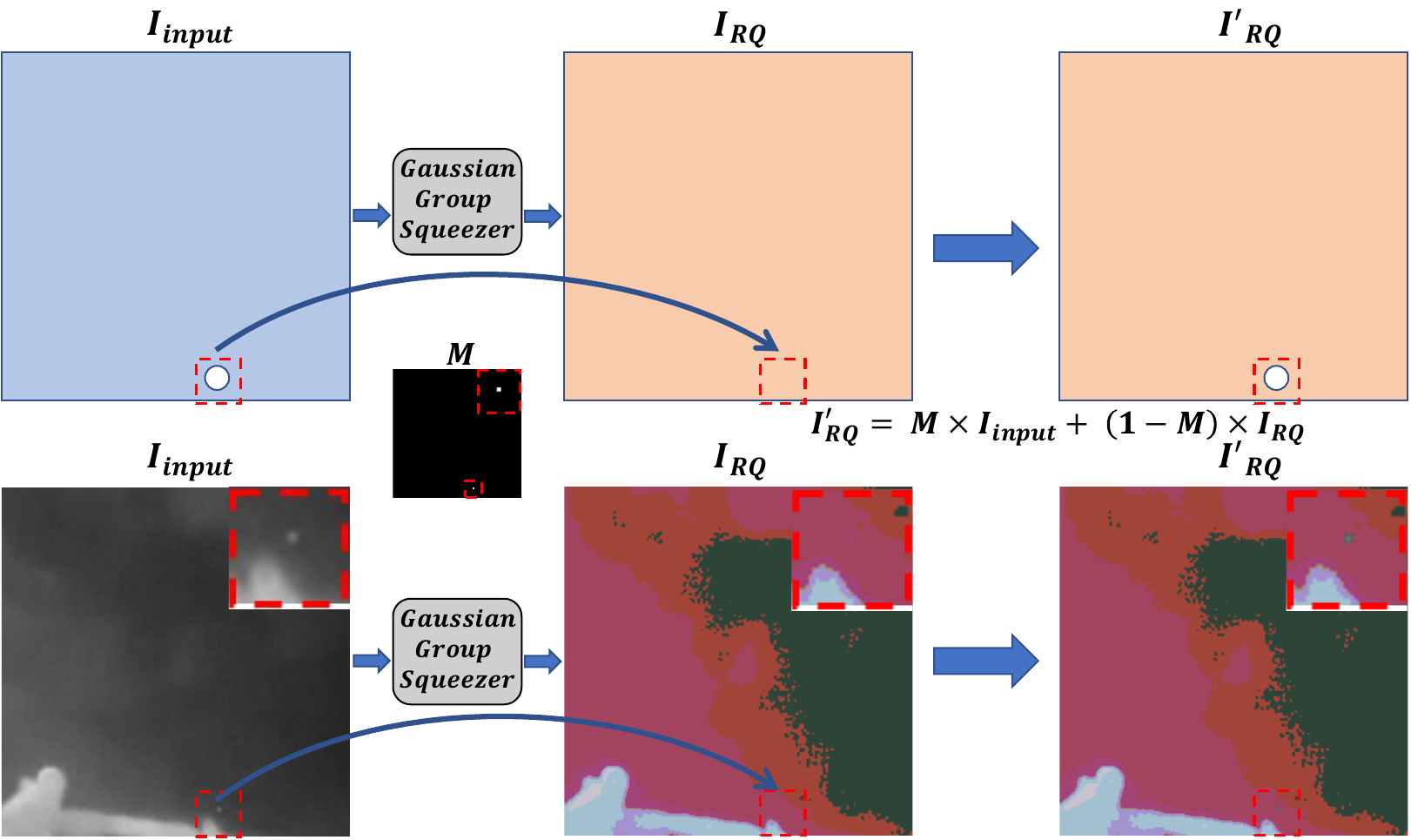}
   \caption{
     Pixel Copy Paste. Quantization operations are applied solely to the background of the infrared image, while the pixels of the small targets remain unquantized, such as $I'_{RQ}$.
}
   \label{fig:fig4}
   \vspace{-5mm}
\end{figure}

\begin{figure*}[t]
  \centering
  \includegraphics[width=0.80\linewidth]{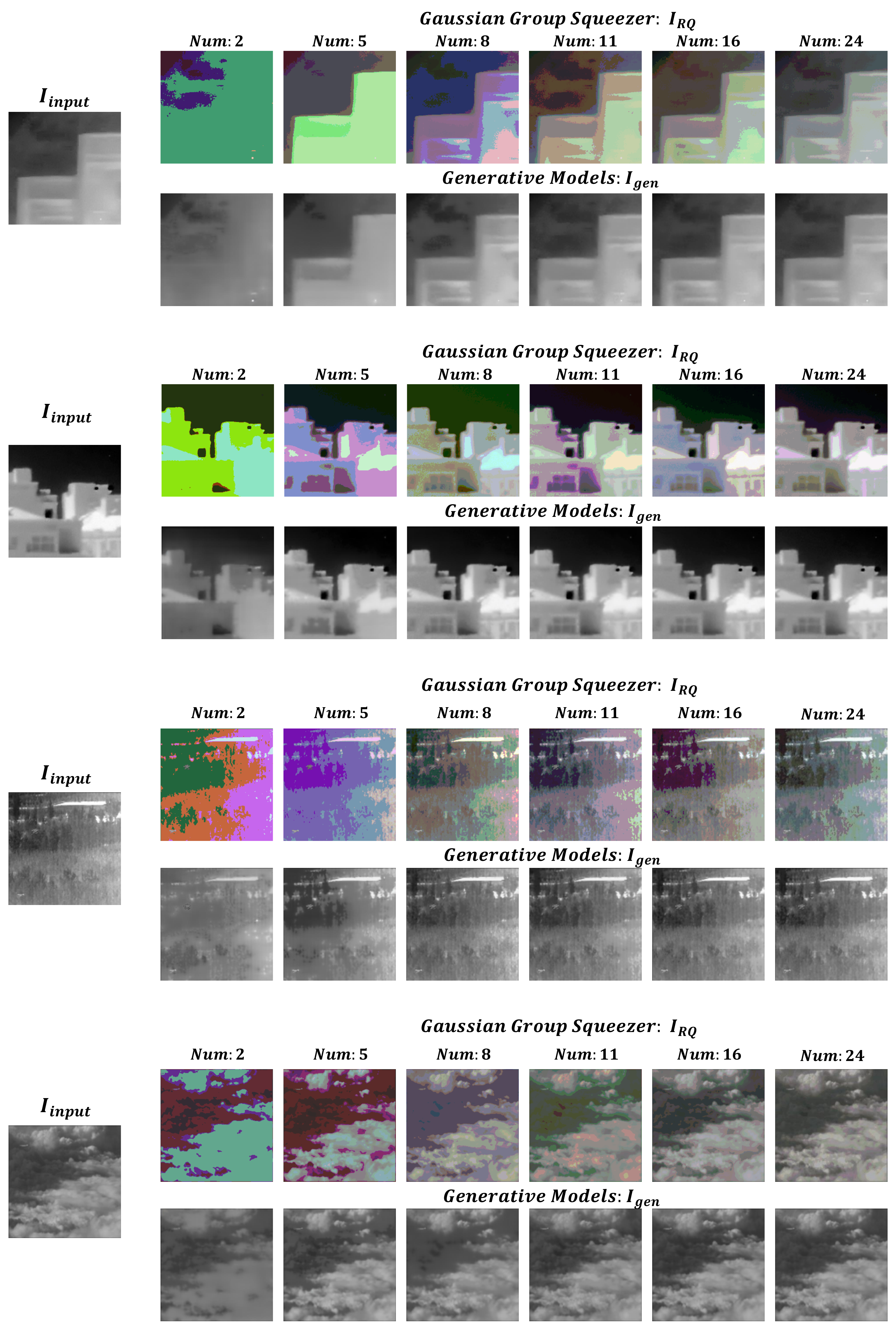}
   \caption{
     Infrared image $I_{gen}$ is generated using diffusion-based generative models. In the training and inference stages, the input image is quantized by different Gaussian parameterized interval. To this end, we generate \emph{agnoistic yet consistent} infrared data.
}
   \label{fig:fig6}
   \vspace{-5mm}
\end{figure*}

\begin{figure}[ht]
  \centering
  \includegraphics[width=0.95\linewidth]{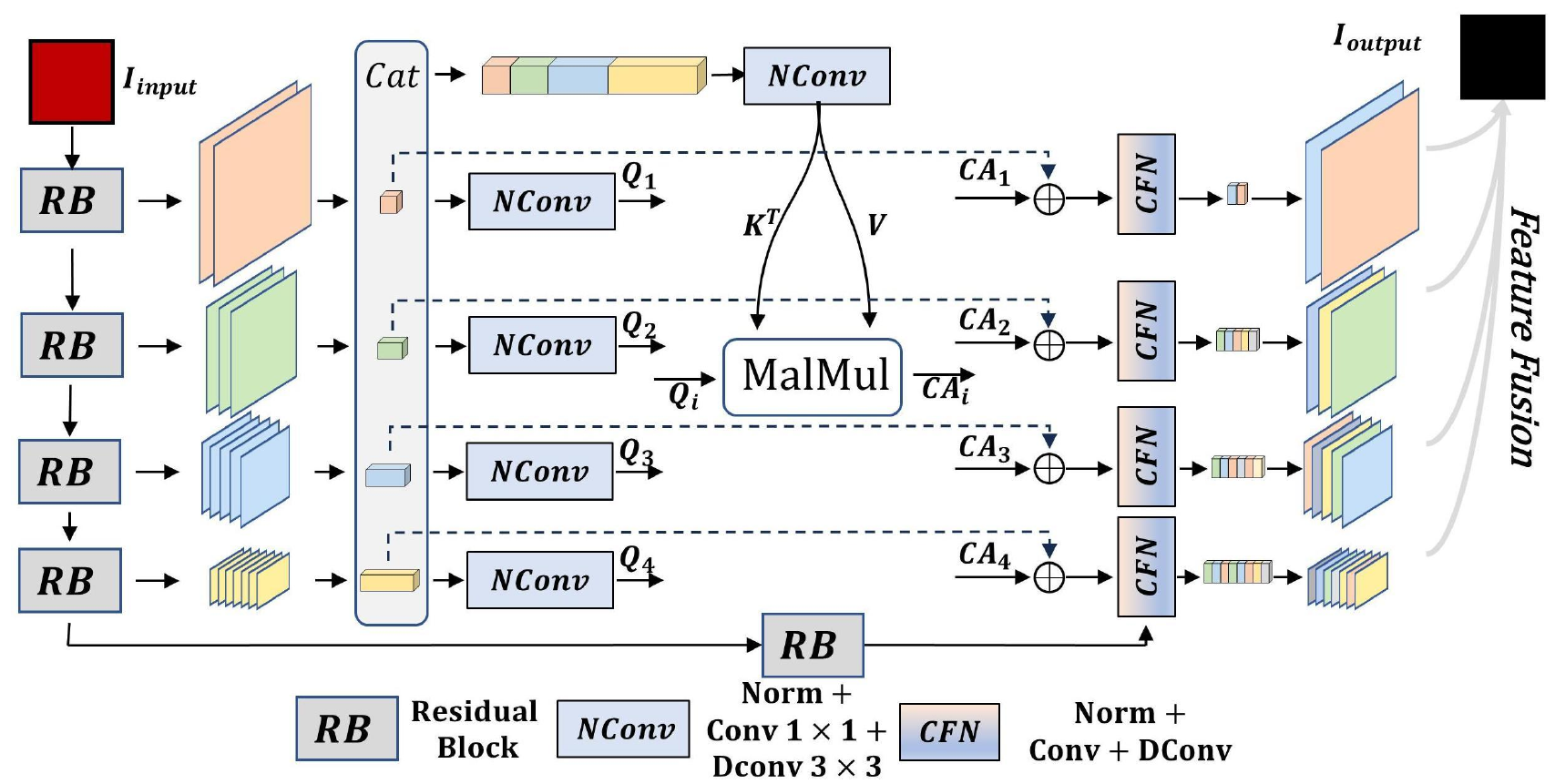}
   \caption{
     Detection framework. We input the image $I_{gen}$ generated by generative modules as an augmentation sample into the detection network for training. 
}
   \label{fig:fig7}
   \vspace{-5mm}
\end{figure}

\subsection{Gaussian Group Squeezer}
Gaussian group squeezer compresses the image at varying degrees by sampling the parameter $Num$ from a Gaussian distribution and integrating it with a diffusion model to generate agnostic enhanced samples. The squeezer discretizes continuous pixel values, thereby reducing data complexity. Different numbers of intervals correspond to varying levels of quantization. The upper part of \autoref{fig:fig3} illustrates the quantization of the image using three intervals for illustration. Specifically, the non-uniform squeezer with randomly assigned interval sizes. It includes a set of non-overlapping intervals $S_i={[a_i, a_{i+1})}$, a corresponding set of replicated values $y_i$, and a parameter $Num$ representing the number of intervals.
\vspace{-3mm}
\begin{equation}
  a_0, a_1, ..., a_{n-1} =\text{sort}(a'_0, a'_1, ..., a'_{n-1})
  \label{eq:eq1}
  \vspace{-5mm}
\end{equation}
\vspace{-8mm}
\begin{equation}
  a'_i = U(\text {min}(x),\text {max}(x)),i=0,1,...,Num-1,
  \label{eq:eq2}
\end{equation}
where $U$ denotes random sampling within the interval, and $\left [ min(x), max(x)\right ] $ represents the minimum/maximum pixel value for each channel of image $x$.
\vspace{-2mm}
\begin{equation}
  y_i=U(a_i,a_{i+1}),i=0,1,...,Num-1,
  \label{eq:eq3}
  \vspace{-3mm}
\end{equation}
$y_i$ is randomly drawn from $U(a_i, a_{i+1})$, replacing the pixels $x$ within that interval to generate the quantized image. As shown in \autoref{fig:fig6}, a larger quantization parameter retains more image information, potentially leading to overfitting during generative model training. Smaller quantization parameters result in a greater loss of image details, hindering model convergence and generalization. To address this effect, we sample quantization parameters from a Gaussian distribution to capture more moderately quantized images with a zero-mean, thereby reducing extreme quantization instances and ensuring balanced sample sizes. By sampling distinct quantization parameters, images with varying compression levels are generated. These quantized images are utilized as inputs for the training and testing of generative models to generate substantial target domain data.

As direct quantization of infrared small-target images often results in the loss of small targets, hindering the reconstruction of pixels around these small targets by generative models. During quantization, the background of the image undergoes the quantization process, while the small target pixels remain unchanged.

\vspace{-5mm}
\begin{equation}
    I_{RQ} = \begin{cases}
    y_i & \text{if } M_{j,k} \neq 1 \text{ and } I_{j,k} \subset [a_i, a_{i+1}) \\
    I_{j,k} & \text{otherwise}
    \end{cases}
\label{eq:eq4}
\end{equation}
Here, $I_{j,k}$ denotes the original pixel value at position $(j, k)$, and $M_{j,k} \in \{0,1\}$ is a binary mask indicating whether the pixel belongs to a small target ($M_{j,k}=1$) or background ($M_{j,k}=0$). $y_i$ is the representative quantized value for the interval $[a_i, a_{i+1})$. Quantization is applied only to background pixels ($M_{j,k} \ne 1$), while small-target pixels are preserved to maintain detail integrity. In \autoref{fig:fig4}, quantizing the entire image leads to the loss of small targets, whereas quantizing only the background while preserving small-target pixels allows the model to more effectively encode the surrounding pixels. Due to the limited number of small-target pixels, which are often lost during image reconstruction, we copy the small-target pixels from the original image to the newly generated image.
\subsection{Two-stage Generative Models}
\begin{figure*}[ht]
  \centering
  \includegraphics[width=0.9\linewidth]{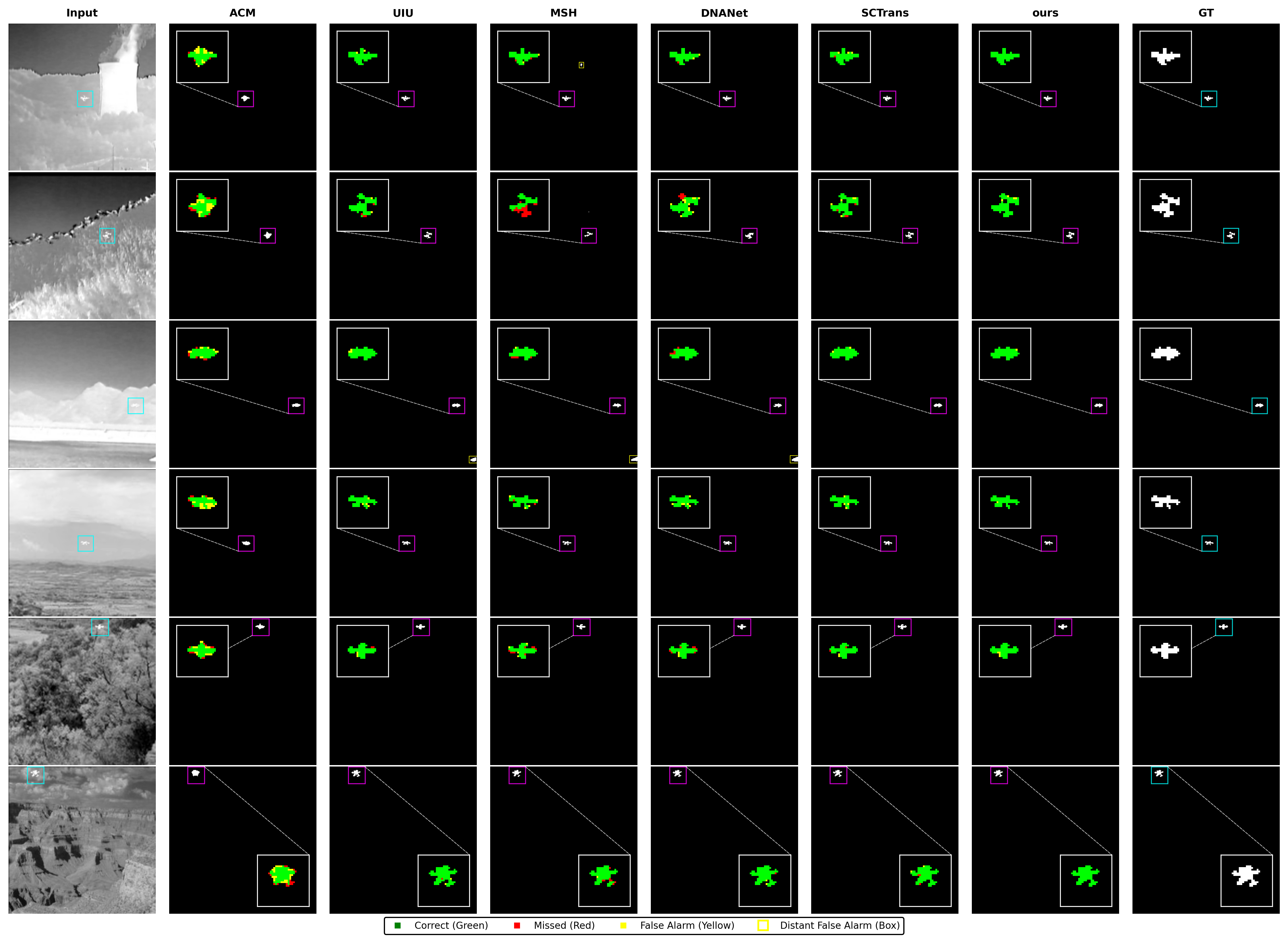}
   \caption{Visualization comparison on the NUDT-SIRST~\citep{li2022dense} and SIRST~\citep{Dai2021} datasets. The outputs generated by our method are more consistent with ground truth labels. In the zoomed-in regions, we use color-coded highlights: \textcolor{green}{green} indicates correctly predicted target pixels, \textcolor{red}{red} marks missed targets (false negatives), and \textcolor{yellow}{yellow} denotes false positives not present in the ground truth.
}
   \label{fig:fig8}
\end{figure*}

\begin{figure*}[t]
  \centering
  \includegraphics[width=0.85\linewidth]{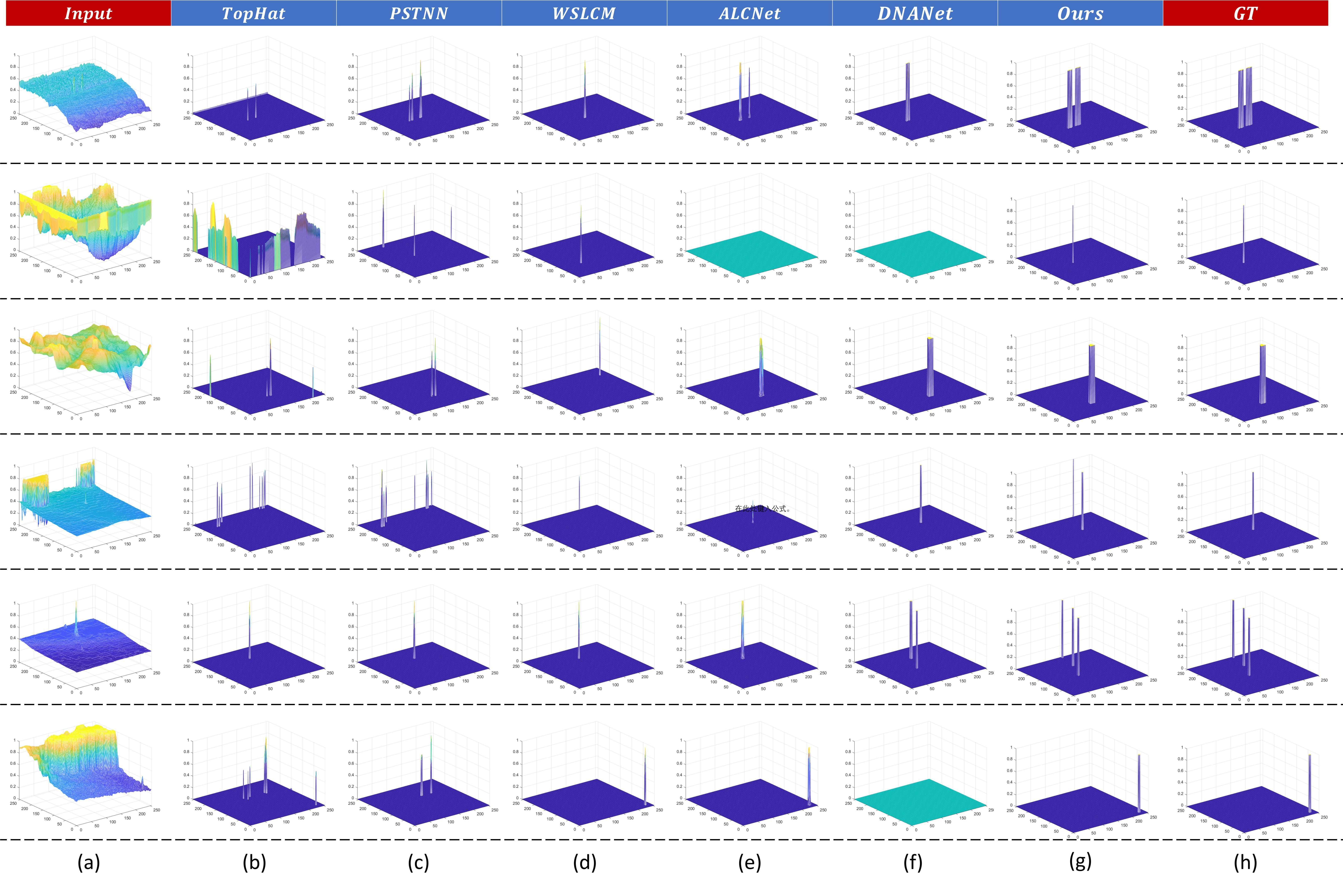}
   \caption{3D visualization results of different methods.
}
   \label{fig:fig3d}
\end{figure*}

Coarse-rebuilding stage aims to learn the mapping from quantized compressed images to real-world representations. By providing the trained model with images at varying levels of quantization, target images with diverse levels of detail are produced. As shown in \autoref{fig:fig5}, where quantization intervals and replication values are randomly selected for each image using the Gaussian group squeezer to enhance the diversity of the training data. Quantized images undergo shallow feature extraction using 3$\times$3 convolutions, and then deeper features are extracted by Residual Swin Transformer Blocks (RSTB) \citep{liu2021swin}. Shallow and deep features are fused and upsampled to the original resolution through three stages of interpolation. Each interpolation stage is followed by a convolutional layer and a Leaky ReLU activation layer. Finally, the coarse-rebuilding process is optimized by minimizing the pixel loss between the original image $I_{input}$ and the generated image $I_{Cr}$ using the following loss function:

\vspace{-5mm}
\begin{equation}
   I_{Cr} = Cr(I_{RQ}), \quad \mathcal L_2 = \left\| I_{Cr} - I_{input} \right\|^2_2
\label{eq:eq5}
\end{equation}
$Cr(\cdot)$ denotes the coarse-rebuilding model applied to the quantized input $I_{RQ}$. The pixel-wise L2 loss $\mathcal{L}_2$ is used to align the generated image $I_{Cr}$ with the ground truth $I_{input}$. This loss encourages structural consistency between the reconstructed and original infrared images. Although the Coarse-rebuilding processed image $I_{Cr}$ shows improvements in quantized pixel quality and distribution alignment, it still falls short of the real-world representations. We feed $I_{Cr}$ into the diffusion stage, allowing it to learn from a large volume of raw infrared small target data. 

In the Diffusion Stage, we employ the Latent Diffusion Model (LDM) \citep{rombach2022high} to generate a priori information by progressively denoising data within the latent representation space and subsequently decoding it into a complete image. The LDM is based on the autoender architecture, consisting of an encoder $\mathcal{E}$ and a decoder $\mathcal{D}$. The input image $I$ is first compressed into a latent representation ${z}$ by the encoder $\mathcal{E}$. At each time step ${t}$, Gaussian noise with variance $\beta_t \in (0, 1)$ is added to generate the noisy latent image ${z}_t$.

\begin{equation}
{z}_t=\sqrt{\bar\alpha_t}z+\sqrt{1-\bar\alpha_t}\epsilon
\label{eq:eq6}
\end{equation}
where $\epsilon \sim \mathcal{N}(0, \mathrm{I})$, and $\bar{\alpha}_t=\prod_{s=1}^t \alpha_s$. As ${t}$ becomes sufficiently large, ${z}_t$ approximates a standard Gaussian distribution. The noise prediction network $\epsilon_\theta$ is trained as a denoising model and optimized using the loss function $\mathcal{L}_{ldm}$:

\vspace{-3mm}
\begin{equation}
\mathcal L_{ldm} = \mathbb{E}_{z,t,c,\epsilon} \left[ \left\| \epsilon - \epsilon_{\theta} \left( {z}_t, c, t \right) \right\|^2_2 \right]
\label{eq:eq7}
\end{equation}
where ${c}$ denotes a conditional variable and $\epsilon$ is sampled from standard Gaussian noise.

To enable the diffusion model to generate high-quality images of the target domain, we fine-tuned it using an infrared small-target dataset. As illustrated in \autoref{fig:fig5}, the image $I_{Cr}$ is encoded into the latent space variable ${z}_0 = \mathcal{E}(I_{Cr})$ by the encoder $\mathcal{E}$ during the training phase. Gaussian noise is then gradually added to obtain ${z}_t$, which approximates a Gaussian distribution. A new latent representation ${z}'_0$ is then generated through ${t}$ resampling steps and decoded by $\mathcal{D}$ to produce the resampled image $I_{gen}$. The loss function $\mathcal{L}_{resample}$ is employed to minimize the difference between the generated image and the realistic image.

\vspace{-3mm}
\begin{equation}
\mathcal L_{resample}=\vert\vert\ {z}'_0-{z}_0\vert\vert^2_2
\label{eq:eq8}
\end{equation}

\begin{table*}[ht]\rmfamily
  \centering
  \setlength{\tabcolsep}{4pt} 
  \caption{Comparison of $IoU(\times 10^{2})$, $Pd(\times 10^{2})$, $Fa(\times 10^{6})$, $Precision(\times 10^{2})$, and $F1(\times 10^{2})$ for Full-Scale and Few-Shot Scenarios on the NUDT-SIRST \citep{li2022dense} Dataset Using Different Methods. The training set consisted of 100\%, 30\%, and 10\% of the images. Larger $IoU$, $P_d$, $Precision$ and $F1$ values, and smaller $F_a$ values, indicate better performance. The best and second-best results are highlighted in red and blue. It is evident that our method achieved the highest performance, demonstrating its effectiveness, particularly in few-shot scenarios.}
  \begin{tabular}{cccccccccccccccc}
    \toprule
    \multirow{2}{*}{Method} & 
    \multicolumn{5}{c}{Full-scale Data (100\%)} & 
    \multicolumn{5}{c}{Few-shot (30\%)} & 
    \multicolumn{5}{c}{Few-shot (10\%)} \\ 
    \cmidrule(lr){2-6} \cmidrule(lr){7-11} \cmidrule(lr){12-16}
    & 
    \makebox[0.03\textwidth][c]{$IoU\uparrow$} & 
    \makebox[0.03\textwidth][c]{$P_d\uparrow$} & 
    \makebox[0.03\textwidth][c]{$F_a\downarrow$} & 
    \makebox[0.03\textwidth][c]{$Prec\uparrow$} & 
    \makebox[0.03\textwidth][c]{$F1\uparrow$} & 
    \makebox[0.03\textwidth][c]{$IoU\uparrow$} & 
    \makebox[0.03\textwidth][c]{$P_d\uparrow$} & 
    \makebox[0.03\textwidth][c]{$F_a\downarrow$} & 
    \makebox[0.03\textwidth][c]{$Prec\uparrow$} & 
    \makebox[0.03\textwidth][c]{$F1\uparrow$} & 
    \makebox[0.03\textwidth][c]{$IoU\uparrow$} & 
    \makebox[0.03\textwidth][c]{$P_d\uparrow$} & 
    \makebox[0.03\textwidth][c]{$F_a\downarrow$} & 
    \makebox[0.03\textwidth][c]{$Prec\uparrow$} & 
    \makebox[0.03\textwidth][c]{$F1\uparrow$} \\
    \midrule
    ACM                     & 67.76 & 96.4  & 13.65 & 76.83 & 80.78 & 64.49 & 95.76 & 15.60 & 74.26 & 78.42 & 53.93 & 91.00 & 64.94  & 62.71 & 70.08 \\
    ALCNet                  & 66.48 & 95.97 & 16.36 & 78.67 & 79.86 & 62.56 & 94.6  & 11.53 & 75.19 & 76.95 & 22.61 & 95.55 & 514.13 & 22.88 & 36.91 \\
    UIUNet                  & 88.39 & \textcolor{blue}{98.62} & 4.87  & 94.26 & 93.84 & 83.7  & 97.67 & 9.65  & 89.56 & 91.13 & \textcolor{blue}{76.68} & \textcolor{blue}{96.61} & 25.94  & 85.82 & \textcolor{blue}{88.70} \\
    MSHNet                  & 77.68 & 95.13 & 10.43 & 90.53 & 87.44 & 75.83 & 95.02 & 21.44 & 87.47 & 86.26 & 64.08 & 91.32 & 68.64  & 80.13 & 78.10 \\
    DNANet                  & 85.21 & 98.51 & 4.06  & 97.26 & 96.85 & 83.17 & 98.09 & 11.42 & 89.11 & 90.82 & 74.07 & 95.66 & \textcolor{blue}{22.45} & \textcolor{blue}{86.16} & 85.11 \\
    SCTrans             & \textcolor{blue}{94.09} & \textcolor{red}{99.04} & \textcolor{blue}{3.95}  & \textcolor{blue}{97.67} & \textcolor{blue}{96.96} & \textcolor{blue}{90.63} & \textcolor{blue}{98.83} & \textcolor{blue}{6.94}  & \textcolor{blue}{95.54} & \textcolor{blue}{95.09} & 71.64 & 95.76 & 27.02  & 84.70 & 83.47 \\ 
    \hline \hline
    \textbf{Ours}           & \textbf{\textcolor{red}{95.37}} & \textbf{\textcolor{red}{99.04}} & \textbf{\textcolor{red}{0.80}} & \textbf{\textcolor{red}{98.06}} & \textbf{\textcolor{red}{97.63}} & \textbf{\textcolor{red}{93.39}} & \textbf{\textcolor{red}{98.94}} & \textbf{\textcolor{red}{2.04}} &
    \textbf{\textcolor{red}{97.17}} &
    \textbf{\textcolor{red}{96.59}} &
    \textbf{\textcolor{red}{86.04}} & \textbf{\textcolor{red}{97.67}} & \textbf{\textcolor{red}{14.13}} &
    \textbf{\textcolor{red}{93.94}} &
    \textbf{\textcolor{red}{92.49}} \\
    \bottomrule
  \end{tabular}
  \label{tab:table2}
  \vspace{-3mm}
\end{table*}

\begin{table*}[ht]\rmfamily
  \centering
  \setlength{\tabcolsep}{4pt} 
  \caption{Comparison of $IoU(\times 10^{2})$, $P_d(\times 10^{2})$, $F_a(\times 10^{6})$, $Precision(\times 10^{2})$, and $F1(\times 10^{2})$ for Full-Scale and Few-Shot Scenarios on the SIRST \citep{Dai2021} Dataset Using Different Methods. The training set consisted of 100\%, 30\%, and 10\% of the images. Larger $IoU$, $P_d$, $Precision$ and $F1$ values, and smaller $F_a$ values, indicate better performance. The best and second-best results are highlighted in red and blue. It is evident that our method achieved the highest performance, demonstrating its effectiveness, particularly in few-shot scenarios.}
  \begin{tabular}{cccccccccccccccc}
    \toprule
    \multirow{2}{*}{Method} & 
    \multicolumn{5}{c}{Full-scale Data (100\%)} & 
    \multicolumn{5}{c}{Few-shot (30\%)} & 
    \multicolumn{5}{c}{Few-shot (10\%)} \\ 
    \cmidrule(lr){2-6} \cmidrule(lr){7-11} \cmidrule(lr){12-16}
    & 
    \makebox[0.03\textwidth][c]{$IoU\uparrow$} & 
    \makebox[0.03\textwidth][c]{$P_d\uparrow$} & 
    \makebox[0.03\textwidth][c]{$F_a\downarrow$} & 
    \makebox[0.03\textwidth][c]{$Prec\uparrow$} & 
    \makebox[0.03\textwidth][c]{$F1\uparrow$} & 
    \makebox[0.03\textwidth][c]{$IoU\uparrow$} & 
    \makebox[0.03\textwidth][c]{$P_d\uparrow$} & 
    \makebox[0.03\textwidth][c]{$F_a\downarrow$} & 
    \makebox[0.03\textwidth][c]{$Prec\uparrow$} & 
    \makebox[0.03\textwidth][c]{$F1\uparrow$} & 
    \makebox[0.03\textwidth][c]{$IoU\uparrow$} & 
    \makebox[0.03\textwidth][c]{$P_d\uparrow$} & 
    \makebox[0.03\textwidth][c]{$F_a\downarrow$} & 
    \makebox[0.03\textwidth][c]{$Prec\uparrow$} & 
    \makebox[0.03\textwidth][c]{$F1\uparrow$} \\
    \midrule
    ACM                     & 66.14 & 91.63 & 24.83 & 82.68 & 80.38 & 62.91 & 91.25 & 33.20 & 76.64 & 77.96 & 20.84 & 85.17 & 657.5  & 21.79 & 34.80 \\
    ALCNet                  & 46.58 & 93.91 & 121.35 & 22.26 & 36.30 & 33.76 & 93.15 & 263.29 & 18.29 & 30.80 & 12.72 & 91.63 & 2192   & 6.93 & 12.95 \\
    UIUNet                  & 76.08 & 92.77 & \textcolor{blue}{13.44} & \textbf{\textcolor{red}{92.53}} & 87.24 & 73.61 & 92.01 & \textbf{\textcolor{red}{15.50}} & \textbf{\textcolor{red}{93.25}} & 85.62 & \textcolor{blue}{62.52} & 91.63 & \textbf{\textcolor{red}{54.81}} & \textbf{\textcolor{red}{83.20}} & 77.73 \\
    MSHNet                  & 73.02 & 95.43 & 19.89 & 79.69 & 79.08 & 69.66 & 94.67 & 46.13 & 75.44 & 76.26 & 59.43 & \textbf{\textcolor{red}{93.91}} & 116.36 & 67.55 & 73.27 \\
    DNANet                  & \textcolor{blue}{76.48} & \textcolor{blue}{96.57} & 26.47 & 88.50 & \textcolor{blue}{87.44} & \textcolor{blue}{74.56} & 93.15 & 18.65 & \textcolor{blue}{89.58} & 86.04 & 56.49 & 88.97 & 143.58 & 73.96 & 72.90 \\
    SCTrans             & 76.00 & 95.81 & 19.34 & 88.43 & 87.17& 69.42 & \textcolor{blue}{95.43} & 63.73 & 89.26 & \textcolor{blue}{88.38} & 59.74 & 90.87 & 94.87  & \textcolor{blue}{83.06} & \textcolor{blue}{79.72} \\ 
    \hline \hline
    \textbf{Ours}           & \textbf{\textcolor{red}{78.88}} & \textbf{\textcolor{red}{96.19}} & \textbf{\textcolor{red}{12.69}} & \textcolor{blue}{89.29} & \textbf{\textcolor{red}{88.98}} & \textbf{\textcolor{red}{78.19}} & \textbf{\textcolor{red}{96.19}} & \textcolor{blue}{16.67} & 
    88.44 & 
    \textbf{\textcolor{red}{88.55}} & \textbf{\textcolor{red}{67.96}} & \textcolor{blue}{92.77} & \textcolor{blue}{80.74} & 
    81.83 & 
    \textbf{\textcolor{red}{81.71}} \\
    \bottomrule
  \end{tabular}
  \label{tab:table6}
  \vspace{-3mm}
\end{table*}

By minimizing the deviation between true and predicted noise, the model learns the data distribution of infrared images, comprehends background semantics and small-target features, and incorporates real-world and diverse knowledge into the images. 

\emph{In inference starge}, we applied the cross-peak sampling strategy. Specifically, during both training and inference, different sets of Gaussian group squeezer parameters were sampled from the same Gaussian distribution to enhance visual diversity. Unlike traditional data augmentation methods, it preserves structural features during the re-generation phase. With the diffusion prior, our method significantly enhances the realism of the generated data, providing more representative samples for discriminative learning. As shown in \autoref{fig:fig6}, $I_{gen}$ is subsequently used for detection network training. The generated images exhibit richer textures and patterns due to the distinct parameter sets of the Gaussian group squeezer during training and inference. 

\subsection{Detection}

To validate the effectiveness of the generated dataset, we combine the generated data $I_{gen}$ with $I_{input}$ for training. As shown in \autoref{fig:fig7}, features are extracted using a ResNet block, followed by enhancement through the spatial attention modules. Finally, features at different scales are fused to generate a robust feature map for predicting $I_{output}$. Both the ground truth $I_{label}$ and the predicted output $I_{output}$ are segmentation maps, and their discrepancy is minimized using the loss function $\mathcal{L}_{IOU}$, defined as follows:
\vspace{-2mm}
\begin{equation}
    \mathcal L_{IOU}=1-\frac{I_{output} \cdot I_{label}+a}{I_{output} + I_{label}-I_{output} \cdot I_{label}+a}
\label{eq:eq9}
\end{equation}

\section{Experiment}
\subsection{Evaluation Metrics}
Conventional pixel-level metrics (IoU, Precision, Recall) adopted by CNN-based works \citep{dai2021asymmetric} \citep{dai2021attentional} \citep{wang2019miss} exhibit two critical limitations for infrared small targets. First, their extreme boundary sensitivity leads to disproportionate penalties: a single misclassified pixel in a 3×3 target reduces Precision by 11.1\%. Second, they misalign with military requirements that prioritize target detection confidence over precise shape delineation. To address these issues, our evaluation framework synergizes tactical and algorithmic metrics:Target-level criteria: $P_d$ for detection rate (critical in early-warning systems) and $F_a$ quantifying false alarms per 10³ pixels.Enhanced pixel metrics: IoU for shape fidelity; F1-score balancing Recall (detection completeness) and Precision (false alarm resistance).

\subsection{Implementation Details}
To demonstrate the effectiveness of our method, we selected SIRST \citep{Dai2021} and NUDT-SIRST \citep{li2022dense} datasets for the experiment. The dataset was split, with 50\% allocated for training and 50\% for testing. The full-scale scenario utilized 100\% of the training data, whereas 30\% and 10\% of the data were used to represent few-shot scenarios at varying scales, allowing validation of the effectiveness of the method under extreme challenges.



During the training phase, the Gaussian group squeezer generated quantization parameters by sampling from a Gaussian distribution with a mean of 17 and a variance of 4. 
The batch size of the Coarse-Rebuilding Stage was set to 3, with 25,000 training steps and a learning rate of 0.001. We borrowed the training method from Diffbir \citep{lin2023diffbir}, loading the diffusion module pre-trained on ImageNet \citep{Deng2009} as the model for the Diffusion Stage. Quantized recovered images were used for fine-tuning, with the batch size, number of epochs, and learning rate set to 1, 20, and 0.01, respectively. During the inference stage, a different set of quantized intervals was sampled. Samples from NUDT-SIRST \citep{li2022dense} and SIRST \citep{Dai2021} datasets were quantized, and the Diffusion Stage utilized DDPM jump sampling with a time step of 50. We generate the same amount of data alongside the original training data for baseline training. The batch size of the detection network was set to 4, the learning rate to 0.05, and the number of epochs to 1000.

\subsection{Comparisons}

We selected several representative methods for comparison, including ACMNet \citep{dai2021asymmetric}, ALCNet \citep{dai2021attentional}, UIU-Net \citep{wu2022uiu}, DNANet \citep{li2022dense}, MSHNet \citep{liu2024infrared} and SCTransNet \citep{10486932}. we retrained all methods on the NUDT-SIRST \citep{li2022dense} and SIRST \citep{Dai2021} datasets following the configurations specified in their respective papers.

As presented in \autoref{tab:table2} and \autoref{tab:table6}, we employed the generated datasets for baseline network training and compared the test results against current state-of-the-art (SOTA) detection networks. Compared to all SOTA methods, our model achieved superior performance across Intersection over Union ($IoU$), probability of detection ($P_d$), and false alarm rate ($F_a$) metrics for both the NUDT-SIRST \citep{li2022dense} and SIRST \citep{Dai2021} datasets. The generated samples significantly improve the performance, proving the reliability of our method. Furthermore, in the few-shot scenario, our method exhibited minimal performance degradation under both full and limited sample quantities. Notably, the performance in the few-shot (30\%) condition of SIRST \citep{Dai2021} outperforms the performance of the SOTA model in the full-scale scenario.

We compared state-of-the-art methods with our method on SIRST \citep{Dai2021} and NUDT-SIRST \citep{li2022dense}. And three metrics, IoU, Pd and Fa
, were used for evaluation. As shown in \autoref{fig:fig8} and \autoref{fig:fig3d}, due to the limited number of target pixels in the image, zoomed-in views were provided for detailed observation. DNA-Net \citep{li2022dense} suffers from leakage and false detection, while UIU-Net \citep{wu2022uiu} and SCTansNet \citep{10486932} fail to accurately detect the target contour in some difficult samples although they reduce false predictions. Detection models trained using samples generated by Generative Models perform well in handling the prediction of difficult samples. This indicates that the data generated by the proposed method in this paper can effectively help the model learn the data distribution and features and provide more challenging samples. Compared with other SOTA methods, our method performs better in $P_d$, $F_a$, and $IoU$ metrics, which significantly improves the accuracy and robustness of ISTD.

We compare common Data augmentation methods with our method to help improve the performance of detection models.\autoref{tab:table7} presents the effects of various data enhancement methods on model performance on the SIRST \citep{Dai2021} dataset. Our method demonstrates superior performance in terms of $IoU$, $P_d$, and $F_a$ metrics. These results indicate that traditional data augmentation techniques can adversely affect the detection of small, making them even harder to identify. In contrast, our proposed enhancement method significantly improves the detection model ability and reduces the false alarm rate.

In \autoref{tab:realtime_efficiency} Our detection model competitive efficiency (10.11 GFLOPs). Diffusion-based augmentation operates strictly offline during training, incurring zero inference overhead. This computational decoupling aligns with data preprocessing optimization principles \citep{wang2021advances}.

\begin{table}[h]\rmfamily
  \centering
  \tabfob
  \caption{The quantitative comparison of our model with various data augmentation techniques on the SIRST \citep{Dai2021} dataset.}
  \begin{tabular}{ccccccccc}
    \toprule
    \multirow{2}{*}{Method} & \multicolumn{3}{c}{Metrics} \\ \cmidrule(r){2-4}
                          & $IoU\uparrow$ & $P_d\uparrow$ & $F_a\downarrow$ \\ 
    \midrule
    Mixup \citep{mixup}                  & 76.89        & 94.67        & 24.97          \\
    CutOut \citep{cutout}                 & 77.57        & \textbf{\textcolor{red}{96.19}}        & 28.40         \\
    CutMix \citep{cutmix}                 & 71.69        & 92.77        & 25.31 \\
    Mosaic \citep{bochkovskiy2020yolov4}                 & 76.10        & 95.81        & 30.59          \\ \hline
    \textbf{Ours}          & \textbf{\textcolor{red}{78.88}} & \textbf{\textcolor{red}{96.19}} & \textbf{\textcolor{red}{12.69}} \\
    \bottomrule
  \end{tabular}
  \label{tab:table7}
\end{table}

\begin{table}[t]\rmfamily
\centering
 \tabfoa
\caption{Real-time detection performance comparison on standard benchmarks. 
Our method achieves the best IoU while maintaining a favorable balance between model size and computational cost.}
\begin{tabular}{lccc}
\toprule
\textbf{Method} & \textbf{IoU (\%)} $\uparrow$ & \textbf{Params (M)} $\downarrow$ & \textbf{FLOPs (G)} $\downarrow$ \\
\midrule
ACM       & 67.76 & \textbf{0.39} & 0.40 \\
ALCNet  & 66.48 & 0.42 & \textbf{0.37} \\
UIUNet & 88.39 & 50.54 & 54.42 \\
DNANet & 85.21 & 4.69 & 14.26 \\
\textbf{Ours}            & \textbf{95.37} & 11.19 & 10.11 \\
\bottomrule
\end{tabular}
\label{tab:realtime_efficiency}
\vspace{-3mm}
\end{table}

\subsection{Ablations}

\begin{table}[t]\rmfamily
  \centering
  \setlength{\tabcolsep}{2pt}
  \tabfoa
  \caption{Ablation results on NUDT-SIRST \citep{li2022dense}, showing that both GGS and diffusion modules contribute significantly to performance, especially in reducing false alarms.}
  \begin{tabular}{lccc}
    \toprule
    Method & $IoU(\times 10^{2})\uparrow$ & $P_d(\times 10^{2})\uparrow$ & $F_a(\times 10^{6})\downarrow$ \\
    \midrule
    Baseline         & 94.09 & 99.04 & 3.95 \\
    w/o GGS          & 93.14 & 98.51 & 3.81 \\
    w/o Diff & 93.64 & 98.51 & 4.18 \\
    Ours & \textbf{95.37} & \textbf{99.04} & \textbf{0.80} \\
    \bottomrule
  \end{tabular}
  \label{tab:table GGS}
  \vspace{-3mm}
\end{table}

\begin{table}[t]\rmfamily
  \setlength{\tabcolsep}{2pt}
  \tabfoa
\centering
\caption{Gaussian parameter sensitivity analysis. Different combinations of mean ($\mu$) and standard deviation ($\sigma$) are tested to evaluate their influence on detection performance.}
\begin{tabular}{ccccc}
\toprule
$\mu$ & $\sigma$ & IoU (\%) $\uparrow$ & $P_d$ (\%) $\uparrow$ & $F_a$ ($\times 10^{-3}$) $\downarrow$ \\
\midrule
8   & 4 & 93.26 & 98.94 & 2.71 \\
28  & 4 & 93.33 & 98.83 & 6.31 \\
17  & 2 & 92.99 & 98.73 & 5.53 \\
17  & 8 & 93.19 & 98.94 & 4.48 \\
\textbf{17}  & \textbf{4} & \textbf{95.37} & \textbf{99.04} & \textbf{0.80} \\
\bottomrule
\end{tabular}
\label{tab:gaussian_sensitivity}
\vspace{-3mm}
\end{table}

\begin{table}[t]\rmfamily
\tabfob
\centering
\caption{Performance comparison on the RealScene-ISTD and IRSTD-1K datasets. 
The evaluation includes Intersection over Union (IoU), probability of detection ($P_d$), and false alarm rate ($F_a$). 
Our method achieves more consistent results across real-world and large-scale benchmarks.}
\begin{tabular}{ccccccc}
\toprule
\multirow{2}{*}{\textbf{Method}} & \multicolumn{3}{c}{\textbf{RealScene-ISTD}} & \multicolumn{3}{c}{\textbf{IRSTD-1K}} \\
& IoU  & $P_d$  & $F_a$  
& IoU  & $P_d$  & $F_a$  \\
\midrule
ACM & 36.86 & 88.03 & 948.4 & 61.79 & 87.98 & 21.82 \\
UIUNet & 37.06 & 86.32 & 680.3 & 64.11 & 91.18 & 29.24 \\
DNANet & 36.25 & 83.76 & 529.1 & \textbf{65.47} & 91.45 & \textbf{19.11} \\
SCTrans & 46.67 & 88.88 & 193.12 & 64.22 & 91.18 & 23.24 \\
Ours & \textbf{50.93} & \textbf{90.59} & \textbf{146.16} & 64.23 & \textbf{93.05} & 23.46 \\
\bottomrule
\end{tabular}
\label{tab:real_irstd_results}
\vspace{-3mm}
\end{table}


\begin{table}[t]\rmfamily
  \centering
  \setlength{\tabcolsep}{2pt}
  \tabfoa
  \caption{Comparison of the Effect of Using RSamp and GSamp on Detection Performance Metrics on the NUDT-SIRST \citep{li2022dense} Dataset. 
 }
  \begin{tabular}{cccccccc}
    \toprule
    & $IoU(\times 10^{2})\uparrow$ & $P_d(\times 10^{2})\uparrow$ & $F_a(\times 10^{6})\downarrow$ & \\
    \midrule
    baseline & 94.09 & 99.04 & 3.95 & \\
    w/ RSamp & 93.50 & 98.51 & 1.88 & \\
    w/ GSamp & \textbf{95.37($\textcolor{red}{\uparrow\textbf{1.87\%}}$)} & \textbf{99.04($\textcolor{red}{\uparrow\textbf{0.53\%}}$)} & \textbf{0.80($\textcolor{red}{\uparrow\textbf{1.08\%}}$)} & \\
    \bottomrule
  \end{tabular}
  \label{tab:table4}
    \vspace{-5mm}
\end{table}
\begin{table}[t]\rmfamily
  \centering
  \setlength{\tabcolsep}{1pt}
  \tabfoa
  \caption{Comparison of the Effects of Cr Stage and Diff Stage on Detection Performance Metrics on the SIRST \citep{Dai2021} Dataset. 
 }
  \begin{tabular}{cccccccc}
    \toprule
    & $IoU(\times 10^{2})\uparrow$ & $P_d(\times 10^{2})\uparrow$ & $F_a(\times 10^{6})\downarrow$ & \\
    \midrule
    baseline & 76.00 & 95.81 & 19.34 & \\
    w/ Cr Stage & 78.50 & \textbf{96.19} & 13.24 & \\
    w/ Diff Stage & \textbf{78.88($\textcolor{red}{\uparrow\textbf{0.38\%}}$)} & \textbf{96.19} & \textbf{12.69($\textcolor{red}{\uparrow\textbf{0.55\%}}$)} & \\
    \bottomrule
  \end{tabular}
  \label{tab:table5}
    \vspace{-5mm}
\end{table}
We conducted multiple sets of experiments using the baseline detection network on the SIRST \citep{Dai2021} and NUDT-SIRST \citep{li2022dense} datasets to evaluate the effectiveness of the individual modules of our proposed method.

\textbf{Effectiveness of the Gaussian Group Squeezer (GGS) and Diffusion Module.} 
As shown in \autoref{tab:table GGS}, both the diffusion module and the GGS contribute significantly to performance.
Removing the diffusion stage leads to a noticeable drop in IoU and a substantial increase in false alarms, indicating its importance in maintaining feature fidelity.
Similarly, disabling the GGS reduces detection accuracy and results in more frequent false positives, highlighting its role in enhancing feature diversity and generalization. \autoref{tab:gaussian_sensitivity} demonstrates that our Gaussian parameter setting ($\mu = 17$, $\sigma = 4$) achieves the best balance between accuracy and stability.
Other configurations yield consistently lower performance, confirming the robustness of our chosen parameters.

\textbf{Effects of Gaussian Sampling (GSamp) and Random Sampling (RSamp).} We compared the model performance using GSamp and random sampling strategies for selecting quantization parameters. As shown in \autoref{tab:table4}, the results indicate that the model employing GSamp significantly outperforms the one using random sampling in terms of $IoU$, $P_d$, and $F_a$ metrics. GSamp ensures a reasonable distribution of quantization parameters, thereby generating more representative samples and enhancing the generalization capability.

\textbf{Effects of Coarse-rebuilding Stage (Cr Stage) and Diffusion Stage (Diff Stage).} To assess the effectiveness of Diff Stage, we compared the model enhanced only with the Cr Stage to the model that combined Cr Stage with Diff Stage resampling. The experimental results demonstrate that incorporating the Diff Stage significantly improves detection performance, particularly in detail recovery and feature reconstruction. The Diff Stage introduces additional real-world information through the resampling process, thereby enhancing the small infrared target detection more effectively. The data enhanced by the Diffusion Stage is more realistic and introduces real-world knowledge. We input the augmented results into the baseline model for experiments. And the quantitative results in \autoref{tab:table5} show that the Diffusion Stage improves performance in detecting small targets. This indicates that the augmented samples generated by the Diffusion Stage are diverse and realistic, effectively enhancing the generalization ability of the model.

\textbf{Effectiveness on More Backbones.}
To assess the effectiveness of using Gaussian agnostic representation learning, we utilized these samples for training various detection models. In this section, we apply UIU-Net \citep{wu2022uiu} and DNA-Net \citep{li2022dense} as detection models and compare their performance with and without Gaussian agnostic data ('w/o Gauss' vs. 'w/ Gauss'). As shown in \autoref{tab:table9}, we compare the performance of DNA-Net \citep{li2022dense} models on the SIRST \citep{Dai2021} dataset with and without Gaussian Agnostic data. To highlight the improvement or degradation, we use uparrow $\uparrow$ to indicate the increase in performance ($IoU$, $P_d$, and $F_a$) and downarrow $\downarrow$ to indicate the decrease. The results demonstrate that training with Gaussian agnostic augmentation significantly improves the performance of DNA-Net \citep{li2022dense} in terms of $IoU$ and $F_a$.

As shown in \autoref{tab:table10}, we evaluated the detection performance of the UIU-Net \citep{wu2022uiu} model on the NUDT-SIRST \citep{li2022dense} (10\%)  dataset. The experimental results demonstrate that the detection performance of UIU-Net \citep{wu2022uiu} improves significantly after training with the generated samples, thereby fully validating the effectiveness of these samples.
\begin{table}[t]\rmfamily
  \centering
  \tabfoa
  \setlength{\tabcolsep}{2pt}
  \caption{Quantitative results of DNANet with and without Gaussian agnostic data on SIRST \citep{Dai2021}.}
  \begin{tabular}{ccccccccc}
    \toprule
    \multirow{2}{*}{Method} & \multicolumn{3}{c}{Metrics} \\ \cmidrule(r){2-4}
                          & $IoU\uparrow$ & $P_d\uparrow$ & $F_a\downarrow$ \\ 
    \midrule
    w/o Gauss                 & 76.48        & 96.57        & 26.47         \\
    w/ Gauss                 &\textbf{77.11($\textcolor{red}{\uparrow\textbf{0.63\%}}$)}        &\textbf{94.67($\textcolor{blue}{\downarrow\textbf{1.90\%}}$)}         &\textbf{14.95($\textcolor{red}{\uparrow\textbf{11.52\%}}$)} \\
    \bottomrule
  \end{tabular}
  \label{tab:table9}
\end{table}

\begin{table}[t]\rmfamily
  \centering
    \tabfoa
  \setlength{\tabcolsep}{2pt}
  \caption{Quantitative results of UIUNet \citep{wu2022uiu} with and without Gaussian agnostic data on NUDT-SIRST ($10\%$). It can be seen that the results of ’w/ Gauss’ are significantly improved compared to ’w/o Gauss’.}
  \begin{tabular}{ccccccccc}
    \toprule
    \multirow{2}{*}{Method} & \multicolumn{3}{c}{Metrics} \\ \cmidrule(r){2-4}
                          & $IoU\uparrow$ & $P_d\uparrow$ & $F_a\downarrow$ \\ 
    \midrule
    w/o Gauss                 &76.68         &96.61         &25.94          \\
    w/ Gauss                 &\textbf{79.17($\textcolor{red}{\uparrow\textbf{2.49\%}}$)}         &\textbf{96.61($\textcolor{red}{\uparrow\textbf{0\%}}$)}         &\textbf{8.29($\textcolor{red}{\uparrow\textbf{17.65\%}}$)}   \\
    \bottomrule
  \end{tabular}
  \label{tab:table10}
\end{table}

\begin{table*}[ht]
\centering
\caption{Nomenclature and Symbol Definitions Used in the Paper}
\begin{tabular}{|c|p{6.5cm}|c|p{6.5cm}|}
\hline
\multicolumn{2}{|c|}{\textbf{Acronyms and Terminology}} & \multicolumn{2}{c|}{\textbf{Mathematical Symbols}} \\
\hline
ISTD & Infrared Small Target Detection &
$I_{j,k}$ & Pixel value at position $(j,k)$ in the input image \\
DDPM & Denoising Diffusion Probabilistic Models &
$M_{j,k}$ & Binary mask (1 = target, 0 = background) \\
LDM & Latent Diffusion Model &
$[a_i, a_{i+1})$ & Interval bounds for non-uniform quantization \\
RSTB & Residual Swin Transformer Block &
$y_i$ & Representative quantized value for bin $[a_i, a_{i+1})$ \\
SOTA & State of the Art &
$I_{RQ}$ & Partially quantized image (target preserved) \\
IoU & Intersection over Union &
$I_{\text{input}}$ & Original unmodified input image \\
$P_d$ & Probability of Detection &
$I_{Cr}$ & Output image from coarse-rebuilding module \\
$F_a$ & False Alarm Rate &
$Cr(\cdot)$ & Coarse-rebuilding generative model \\
-- & -- &
$\mu, \sigma$ & Mean and standard deviation of the Gaussian distribution used in GGS ($\mu{=}17$, $\sigma{=}4$) \\
-- & -- &
$\mathcal{L}_2$ & L2 reconstruction loss between $I_{Cr}$ and $I_{\text{input}}$ \\
\hline
\end{tabular}
\end{table*}
The results in Table~\ref{tab:real_irstd_results} further validate the cross-domain generalizability of our method. Despite the differences in sensor types, scene structures, and target characteristics between RealScene-ISTD\citep{lu2025rethinking} and IRSTD-1K\citep{zhang2023isnet}, our approach maintains high accuracy and low false alarm rates, indicating its strong adaptability to real-world infrared environments.

\section{Conclusion}
In this paper, we proposed a novel Gaussian Agnostic Representation Learning framework to address the extreme challenges of infrared small target detection. The proposed Gaussian group squeezer generates diverse training data through non-uniform quantization, while the coarse-rebuilding and diffusion stages enhance data quality by reconstructing target structures and incorporating prior knowledge from diffusion models. Experimental results demonstrate that our method significantly improves detection accuracy in extreme conditions compared to state-of-the-art methods. The effectiveness of our approach highlights its potential for overcoming the limitations of dataset scale in infrared target detection, providing a promising direction for future research.

\bibliographystyle{cas-model2-names}

\bibliography{cas-refs}



\end{document}